\documentclass[lettersize,journal]{IEEEtran}
\usepackage[numbers]{natbib}
\usepackage{graphicx}
\usepackage{enumitem}
\usepackage{setspace}
\usepackage{graphicx}
\usepackage{caption}
\usepackage{subcaption}
\usepackage{multicol}
\usepackage{lmodern}
\usepackage{hyperref}
\usepackage{float}
\usepackage{amsmath,amssymb,amsfonts}
\usepackage{algpseudocode}
\usepackage{algorithm}
\usepackage{array}
\usepackage{textcomp}
\usepackage[dvipsnames]{xcolor}
\algnewcommand\algorithmicinput{\textbf{Input:}}
\algnewcommand\INPUT{\item[\algorithmicinput]}
\algnewcommand\algorithmicoutput{\textbf{Output:}}
\algnewcommand\OUTPUT{\item[\algorithmicoutput]}
\algnewcommand{\algorithmicand}{\textbf{ and }}
\algnewcommand{\algorithmicor}{\textbf{ or }}
\algnewcommand{\OR}{\algorithmicor}
\algnewcommand{\AND}{\algorithmicand}

\DeclareMathOperator*{\argmin}{argmin}
\DeclareMathOperator*{\argmax}{argmax}

\begin{document}

\title{Privacy-Preserving Federated Learning for Fair and Efficient Urban Traffic Optimization}

\author{Rathin Chandra Shit and Sharmila Subudhi
        % <-this % stops a space
\thanks{Rathin Chandra Shit received the Ph.D. degree from IIIT, Bhubaneswar, India. His research interests include Edge AI, Federated Learning, Swarm Robotics, Continual Learning and Few-Shot Learning. Email: rathin088@gmail.com ORCID: 0000-0003-0642-9695}% <-this % stops a space
\thanks{Sharmila Subudhi is with Dept. of CS, Maharaja Sriram Chandra Bhanja Deo University, Baripada, Odisha, India. Email: sharmilasubudhi@ieee.org}
}

%% The paper headers

%% Remember, if you use this you must call \IEEEpubidadjcol in the second
%% column for its text to clear the IEEEpubid mark.

\maketitle

\begin{abstract}
The optimization of urban traffic is threatened by the complexity of achieving a balance between transport efficiency and the maintenance of privacy, as well as the equitable distribution of traffic based on socioeconomically diverse neighborhoods. Current centralized traffic management schemes invade user location privacy and further entrench traffic disparity by offering disadvantaged route suggestions, whereas current federated learning frameworks do not consider fairness constraints in multi-objective traffic settings. This study presents a privacy-preserving federated learning framework, termed FedFair-Traffic, that jointly and simultaneously optimizes travel efficiency, traffic fairness, and differential privacy protection. This is the first attempt to integrate three conflicting objectives to improve urban transportation systems. The proposed methodology enables collaborative learning between related vehicles with data locality by integrating Graph Neural Networks with differential privacy mechanisms ($\epsilon$-privacy guarantees) and Gini coefficient-based fair constraints using multi-objective optimization. The framework uses federated aggregation methods of gradient clipping and noise injection to provide differential privacy and optimize Pareto-efficient solutions for the efficiency-fairness tradeoff. Real-world comprehensive experiments on the METR-LA traffic dataset showed that FedFair-Traffic can reduce the average travel time by 7\% (14.2 minutes) compared with their centralized baselines, promote traffic fairness by 73\% (Gini coefficient, 0.78), and offer high privacy protection (privacy score, 0.8) with an 89\% reduction in communication overhead. These outcomes demonstrate that FedFair-Traffic is a scalable privacy-aware smart city infrastructure with possible use-cases in metropolitan traffic flow control and federated transportation networks.
\end{abstract}

\begin{IEEEkeywords}
Federated learning, traffic optimization, privacy preservation, fairness-aware systems, graph neural networks, differential privacy, connected vehicles, smart cities, multi-objective optimization
\end{IEEEkeywords}

\section{Introduction}
\IEEEPARstart{T}{he} 21st century has witnessed an unprecedented enhancement of Urban Transportation Services (UTS), which has changed how urban traffic is handled and thus assists in attaining in-local mobility trends \cite{mahrez2021,jiang2024smart,savastano2023smart}. The modern traffic control dashboard defines the basic infrastructure of a smart urban transportation system with a large amount of confidential information recording the activities of its users, including the real-time position and behavior of the users on the road \cite{chen2024intelligent,liu2023cloud}. Even though such kind of centralized systems has been demonstrated to be exceptionally efficient in the delivery of increased overall traffic circulation and the ease in decongestion relief, it cannot be compared with the standards of privacy and fair division of traffic delivery. This often causes imbalanced patterns of traffic flow in certain locations where there are a lot of automobile movements in comparison to other locations, causing artificially elevated amounts of traffic \cite{koukounaris2023connected,musa2023sustainable}.

The inherent challenge is that the conflict between system-wide optimization and the rights to privacy of individuals cannot be overlooked since this is often entangled with the socioeconomic implications of traffic routing algorithms. The current centralized approaches employ the following global efficiency criteria of overall travel time within a system and/or fuel consumption, which is probably optimized without paying much concern to the population difference \cite{subbiah2024}. It has been noted that such optimization methods could be systematically harmful to underrepresented neighborhoods and/or low socioeconomic neighborhoods since the optimization algorithms will always attempt to route traffic through these areas at the least possible systemic cost.

Connected/autonomous vehicle evolution gives a possibility to completely redefine the concept of traffic optimization through the distributed mechanism and local cooperative decision-making (Osmecka and Radcliffe, 2024; Smitha, 2025). Connected cars have sophisticated sensors, their communications channels, and their computing opportunities can form a decentralized network that can communicate on traffic and coordinate on the exchange of routing decisions without processing the information to a central point \cite{gao2022v2v,zhang2022vsn,Shit2018a}. Existing distributed methods, however, are highly limited as they either sacrifice privacy by sharing a lot of information with other participants or do not (at least explicitly) consider trade-offs in the fairness of route suggestions when it comes to fundamental fairness issues \cite{zhang2024}.

The threats posed by such a critical trade-off between privacy preservation, political fairness, and performance efficiency present a strong necessity to upgrade the current UTS that will allow a multitasking optimization of traffic efficiency, privacy preservation of individual users, and fair distribution of traffic loads among the different urban communities. The problem is particularly sharp with regard to the example of a smart city with the introduction of Internet of Things (IoT) devices, artificial intelligence, and big data analytics into the urban space. They offer new potential in terms of better cities; however, questions arise regarding surveillance, discrimination, and denudation of privacy \cite{jabbar2024internet,li2023}.

In this study, we introduce a comprehensive privacy-preserving federated learning framework called FedFair-Traffic (FFT). Empirical studies show that in the process of overcoming these multifaceted issues, architecture  makes significant technical contributions and effective implementation strategies. Our model signifies a paradigm transition from conventional centralized optimization to a distributed, privacy-conscious, and fairness-aware approach to urban traffic management. The key contributions of this study are as follows.

\begin{itemize}
	\item A proposed federated graph neural network that can learn complex patterns of traffic efficiently and strictly ensure user privacy using state-of-the-art differential privacy. This makes it impossible for the data of individual users to be derived from the learned model.
	\item A complex and multi-objective optimization model that considers the effectiveness of the UTS as well as fairness in the form of Pareto-optimal solutions. This clearly mediates between alternative goals based on mathematically grounded methodologies.
	\item A novel method of implementation which drastically minimizes the overhead on the communication, even as the system performance remains high due to superior federated learning algorithms and optimal model aggregation methods.
	\item Extensive experimental evaluation on a wide range of real-world datasets showing an improved performance based on top metrics in terms of privacy, fairness, and efficiency, where we statistically analyze the results and compare to the state-of-the-art baselines.
	\item A functional framework that can be used by anyone to implement and make research findings reproducible, and hence, accessible to the research community.
\end{itemize}

The structure of the rest of the paper is given in the following way: Section \ref{sec2} contains a literature review of privacy-preserving systems, fairness in the transportation industry, and the application of federated learning solutions. A technical statement of the problem in mathematical terms, together with structural assumptions and constraints, are stated in section \ref{sec3}. Section \ref{sec4} sets out to suggest the claim of the feasible FedFair-Traffic (FFT) architecture and components both in architecture design and algorithm subparts. Section \ref{sec5} explains the data, layout and discussed the various dimensions of the experiments. Implementation considerations and practical deployment considerations are stated in section \ref{sec6}. Lastly, Section \ref{sec7} draws conclusion to the study discuss several future research directions.

\section{Related Work}\label{sec2}
Privacy is a key component of traffic systems; therefore, the key of traffic management system are due to the vast deployment of a sensing and data harvesting infrastructure \cite{wang2022,aung2023vsn,Shit2018b}. The initial privacy efforts in traffic management involved the use of anonymization procedures, as well as information mixing algorithms, and aimed at hiding people identities in the system without causing potentially any impacts on the utility of the system \cite{zhou2024}. However, these techniques have been discovered to be vulnerable to a number of identification attacks and correlation researches, the privacy of which could undermine the privacy of users \cite{ali2021traffic}. Several higher-utility privacy-enhancing models have been designed in the recent past including, but not limited to, late homomorphic encryption, secure multi-party computation, and differential privacy models \cite{lee2024}.

Maintaining location privacy and enabling effective optimization of the traffic makes the problem of mobility data far more complicated, because both temporal and spatial correlations are used. Although it offers protection to point-based locations, multi-source data fusion through relevant inference methods can be used to recreate detailed user-profiles and their movements. This vulnerability requires an improved form of privacy assurance beyond privatizing data or aggregating data practices.

In addition, differential privacy has become the standard of privacy protection even in recent machine learning systems. These represent a guarantee of privacy loss that is incurred in involving oneself in a given data set. There have been initial efforts to extend these techniques to traffic optimisation systems, although the majority of previous literature has been related to off-line based decision making analytics rather than online based decision making processes \cite{lee2024}.

\subsection{Fairness in Transportation Systems}\label{sec22}

Equality in the load distribution in transportation can be disaggregated into various sectors in regard to demography, space, and availability of resources \cite{koukounaris2023connected}. The conventional models of transportation planning care little about the post-transportation impacts of transportation on communities of society in terms of efficiency indicators, including the sum of the travel time of the system or throughput \cite{zhao2022}.

Sustainability, equality of load distribution, and impacts of transport infrastructure investments on various socioeconomic classes has been widely used as a basis of traffic fairness \cite{musa2023sustainable}. Nevertheless, there has been no study done at the neighborhood level to analyze the spatial fairness effect in real-time optimization systems in traffic \cite{subbiah2024}. The existing research is biased against planning and infrastructure developments, and provides or lacks support to the dynamic traffic management decisions.

Research on algorithmic fairness in the transportation industry has also been investigated to determine how the prejudice in decision-making is being amplified with the implementation of automated decision-making and how such a system can increase the degree of discrimination \cite{kumar2022ev,rani2024review}. This involves studying ride-sharing algorithms and optimization procedures in enhancing the schedule of the advantageous transport structure and the timing of the traffic lights. However, the principle of fairness in minimizing traffic in a connected vehicle network is not fully researched in the field.

As discussed, it is quite evident that measurement and quantification of fairness in transportation systems is a possibly a huge challenge depending on the specific context and the interpretation of various stakeholders because fairness is operational in different ways depending on its definition. A set of widely used fairness measures relates to a distributional measure of fairness, which is based on demographic parity and equal opportunity information in order to assess distributional fairness on the basis of Gini coefficient. Choosing a valid fairness measure and incorporating the fairness measure in the optimization system ought to be selective and application-relevant.

\subsection{Federated Learning in IoT and Vehicular Networks} \label{sec23}

The trend towards federated learning has emerged as a potential method of distributed solutions to machine learning challenges that are imminent in regard to data privacy and communication overheads \cite{sei2025}. The principal theoretical concept of federated learning is that the training of models can be accomplished in a decentralized mode and without the necessity to use a central storage hub \cite{li2023}, which, in fact, is quite close to the requirements of vehicular networks and IoT systems.

Moreover, vehicular networks have been used to apply federated learning to both collaborative sensing and traffic prediction in a bid to consider the possibility of distributed learning in mobile networks \cite{zhang2024,zhou2024,shit2025aipoweredanomalydetectionblockchain}. These studies introduced important understanding about communication behavior, computation loads, and convergence of the algorithm in several vehicular setting.

A common federated learning use case in a UTS is often focused on aggregative statistics and learning historical behavior rather than geo-distributed real- time the route-optimization \cite{chen2025mobility}. Current methods assume relatively stable network graphs and do not work in the dynamics of the vehicular system, whereby nodes may constantly enter or leave the system \cite{noussaiba2023heterogeneous,falahat2023,shit2025multirobottaskallocationhomogeneous}. In addition, it can be considered that the integration of fairness constraints into the federated learning structure can be viewed as an unexplored research question, especially viewed in the context of multi-objective optimization problems.

In vehicular networks, the effectiveness of the federated learning protocol is a matter of concern, especially due to network bandwidth constraints and the need to make decisions in real-time \cite{mesdaghi2024improve,niu2025}. The newer advances to federated learning include the minimization of the communications overhead through models compression, gradient quantization, and by model update \cite{huang2024low}. Nevertheless, these levels of efficiency should be compared with the privacy and fairness requirements of the traffic optimization systems.

\subsection{Graph Neural Networks for Traffic Analysis} \label{sec24}
Graph Neural Networks (GNNs) have been proven to work well in modeling complex spatiotemporal connectivity in transportation networks \cite{khan2024deep,chen2022}. The unique graphical structure of road networks aids the GNN model in predicting traffic and implementing optimization solutions because they naturally handle the connectivity patterns and interdependencies within the road network and road segments \cite{giuliano2023musli}.

The GNN architecture in \cite{zhang2024platoon} introduced the element of an optimized attention mechanism and the ability of time modeling with multi-scale representation learning, thus enhancing the performance of the accuracy of the traffic predictions and rendering them more interpretable. Nonetheless, the majority of known GNN implementations in traffic systems run in a centralized environment and do not consider any privacy or fairness aspects. 

Therefore, we can infer from the aforementioned discussions that using GNNs in federal learning systems may solve problems in terms of model aggregation, exchange of graph structures, and efficiency of communication.
\section{Problem Formulation}\label{sec3}
This section presents the mathematical formulation of the privacy-preserving fair traffic optimization problem and offers a theoretical justification for the proposed \textbf{FedFair-Traffic (FFT)} framework.

\subsection{System Model and Network Representation} \label{sec31}
For example, an urban road network can be modeled as a weighted directed graph, $G = (V, E, W)$. Set $V = \{v_1, v_2, \ldots, v_{|V|}\}$ corresponds to a part of the road or intersection point. $E \subseteq V \times V$ represents the direct connection between two parts of the road. Furthermore, the weight ($W$) is composed of traffic characteristics (e.g., travel time, distance, and congestion level).

The geographical landscape of the urban domain is divided into $K$ disjoint communities or geo-regions $\mathcal{C} = \{C_1, C_2, \ldots, C_K\}$ such that the architectural location and development of each $C_K$ is based on similar socioeconomic conditions, demographic patterns, or even administrative areas. Such partitioning allows the analysis of traffic fairness at the neighborhood level and can be used to inform policy-relevant considerations on equity.

Let $N_t$ be the number of active vehicles available in the network at time $t$. Each can be described as a vehicle ($veh_i \in \{1, 2, \ldots, N_t\}$. The other notations used in the modeling are as follows:

\begin{itemize}
\item The current location of vehicle $veh_i$ in $V$ (i.e., $\ell_i(t) \in V$) expresses the road portion or crossing that the vehicles are currently at.
\item Destination $d_i \in V$, to which the trip is aimed.
\item Local traffic measurements $\mathbf{x}_i(t) \in \mathbb{R}^{n}$ that consist of sensor data like speed, density, flow rates and environmental data.
\item Travel preferences $\mathbf{p}_i \in \mathbb{R}^{m}$ that represent the user-preference in route selection such as the time, fuel used, scenery and landscape, type of road etc.
\end{itemize}
A global representation of the traffic state at time $t$ is given by $\mathbf{S}(t) = \{\mathbf{s}_v(t) : v \in V\}$, where $\mathbf{s}_v(t) \in \mathbb{R}^{d}$ is related to the state of road segment $v$ which includes the traffic density, speed at which vehicles move, queue value, and indicators of accidents.

\subsection{Privacy Requirements and Differential Privacy Framework} \label{sec32}
The protection of privacy and the structure of our FedFair-Traffic (FFT) architecture are mathematically modeled with the high precision of differential privacy, which offers quantifiable assurances of the privacy loss involved in engaging data in learning.

Furthermore, we demand that $(\epsilon, \delta)$-differential privacy be realized during federated learning, so that satisfying the inequality holds (as given in Eq. \eqref{eq1}) output set $S$ for arbitrary neighboring dataset $\mathcal{D}$ and $\mathcal{D}'$ that differ by at most one record.
\begin{equation}\label{eq1}
\Pr[\mathcal{A}(\mathcal{D}) \in S] \leq e^{\epsilon} \Pr[\mathcal{A}(\mathcal{D}') \in S] + \delta
\end{equation}
where $\mathcal{A}$ is a randomized learning algorithm with the privacy parameter $\epsilon > 0$. This characterizes the extent to which the privacy of presidents is to be guarded (the lower the value, the more promises there are to guard the privacy of presidents). In addition, $\delta \geq 0$ is an indicator of the possible failure of the presidential promise of privacy.

The privacy budget $\epsilon$ should be used carefully across $T$ learning rounds with different participants, to ensure that the total privacy loss is acceptable. We used more advanced composition theorems to trace the privacy expenditure across time and iterations of federated learning in the form of Eq. \eqref{eq2}.
\begin{equation}\label{eq2}
\epsilon_{\text{total}} \leq \sqrt{2T \ln(1/\delta)} \epsilon + T\epsilon(\exp(\epsilon) - 1)
\end{equation}

\subsection{Fairness Metrics and Quantification}\label{sec33}
In our model, fairness is defined as the ability to exert traffic loads among geographic regions in a fair manner to prevent a systematic bias in congestion concentration in specific neighborhoods. Several complementary fairness metrics are used by us in order to measure the various facets of distributional equity.

Typically, fairness equality is measured by the Gini coefficient of the traffic load distribution ($G_{traffic}$) in region $C_i$, as expressed in Eq. \eqref{eq3}.
\begin{equation}\label{eq3}
G_{\text{traffic}} = \frac{1}{2K^2\mu} \sum_{i=1}^{K} \sum_{j=1}^{K} |L_i - L_j|
\end{equation}
where $L_i$ is the normalized traffic load in region $C_i$, $K$ is the number of regions, and $\mu = \frac{1}{K}\sum_{i=1}^K L_i$ is the mean traffic load across all the regions. Similarly, the traffic load in region $C_i$ was determined using Eq. \eqref{eq4}.
\begin{equation}\label{eq4}
L_i = \frac{1}{|C_i|} \sum_{v \in C_i} \frac{\text{flow}_v(t)}{\text{capacity}_v}
\end{equation}
Here, $\text{flow}_v(t)$ is the flow of the vehicle on road segment $v$ at a given time $t$ and $\text{capacity}_v$ and is the capacity of road $v$. In addition, we provide a temporal fairness measure that considers the length of traffic inequality in Eq. \eqref{eq5}.
\begin{equation}\label{eq5}
G_{\text{temporal}} = \frac{1}{T} \sum_{t=1}^{T} w_t \cdot G_{\text{traffic}}(t)
\end{equation}
where $w_t$ denotes the time-varying weights that may specify peak hours or other important time intervals.

\subsection{Multi-Objective Optimization Framework}\label{sec34}
The fundamental optimization scenario in the FedFair-Traffic (FFT) architecture is to minimize travel time and maximize traffic fairness, which are generally incompatible objectives. This can be constructed to form a multi-objective optimization issue (see Eq. \eqref{eq6}).

\begin{align}\label{eq6}
\min_{\theta, \mathcal{R}} \quad &\mathbf{F}(\theta, \mathcal{R}) = \begin{bmatrix} f_{\text{time}}(\theta, \mathcal{R}) \\ f_{\text{fairness}}(\theta, \mathcal{R}) \end{bmatrix} \\
\text{subject to:} \quad &(\epsilon, \delta)\text{-differential privacy,} \nonumber \\
&\sum_{r \in \mathcal{R}} x_r \leq \text{capacity constraints,} \nonumber \\
&\text{flow conservation constraints,} \nonumber \\
&\text{and non-negativity constraints} \nonumber
\end{align}
where $f_{\text{time}}(\theta, \mathcal{R})$ is the average travel time loss to be minimized. Similarly, $f_{\text{fairness}}(\theta, \mathcal{R})$ is the degree of fairness that should be maximized. The goal is to determine $\theta$ (the federated learning parameter) in the given route assignments $\mathcal{R}$. The travel time objective is formulated in Eq. \eqref{eq7}.
\begin{equation}\label{eq7}
f_{\text{time}}(\theta, \mathcal{R}) = \frac{1}{N} \sum_{i=1}^{N} \sum_{e \in r_i} w_e(\mathbf{S}(t), \theta)
\end{equation}
where $r_i \in \mathcal{R}$ is the route assigned to vehicle $veh_i$, and $w_e(\mathbf{S}(t), \theta)$ is the predicted travel time on edge $e$ given the current traffic state and model parameters.

Similarly, the consideration of the fairness objective is also combined in time and space, yielding the objective in Eq. \eqref{eq8}.
\begin{equation}\label{eq8}
\begin{split}
f_{\text{fairness}}(\theta, \mathcal{R}) = \alpha_s G_{\text{spatial}}(\mathcal{R}) + \alpha_t G_{\text{temporal}}(\mathcal{R}) + \\
\alpha_d D_{\text{demographic}}(\mathcal{R})
\end{split}
\end{equation}
where $\alpha_s$, $\alpha_t$, and $\alpha_d$ are weighting parameters, and $D_{\text{demographic}}(\mathcal{R})$ captures demographic fairness considerations. To address this multi-objective optimization problem, we applied a weighted scalarisation method to adjust the adaptive weights, as expressed in Eq. \eqref{eq9}.
\begin{equation}\label{eq9}
\min_{\theta, \mathcal{R}} \quad \lambda(t) f_{\text{time}}(\theta, \mathcal{R}) + (1-\lambda(t)) f_{\text{fairness}}(\theta, \mathcal{R})
\end{equation}
where $\lambda(t) \in [0,1]$ is a time-varying weight that can be adjusted based on current traffic conditions, policy priorities, or learned preferences.

\section{FedFair-Traffic Framework}\label{sec4}
In this section, we outline the detailed architecture and algorithmic elements of our FedFair-Traffic (FFT) framework by incorporating the concepts of federated learning, graph neural network (GNN), differential privacy (DP), and fairness-aware optimization.

\subsection{Architecture Overview and System Components} \label{sec41}
In the FFT framework, a distributed architecture is adopted that provides learning and decision-making features to make the process collaborative and to facilitate individual privacy and system-wide fairness. The structure had a four-layered interconnected architecture.

\begin{enumerate}[label=(\roman*)]
\item \textbf{Vehicle Layer}: Local models are maintained and traffic observations are gathered on an individual per-vehicle basis using local computing resources, communications, and sensor capabilities.
\item \textbf{Edge Computing Layer}: Edge servers and Roadside units (RSU) are present in this layer to communicate, conduct intermediate computation and coordinate local vehicle clusters.
\item \textbf{Federated Coordination Layer}: Regional coordination servers are employed in this layer to collectively update the pool model, coordinate privacy-preserving protocols, and organize fairness-aware optimization.
\item \textbf{Global Policy Layer}: This layer focus on a city-wide policy management system, whose role is to be in control of fairness objectives, privacy requirements, and system-wide optimization goals.
\end{enumerate}

\begin{figure*}[!htbp]
\centering
\includegraphics[width=1\textwidth]{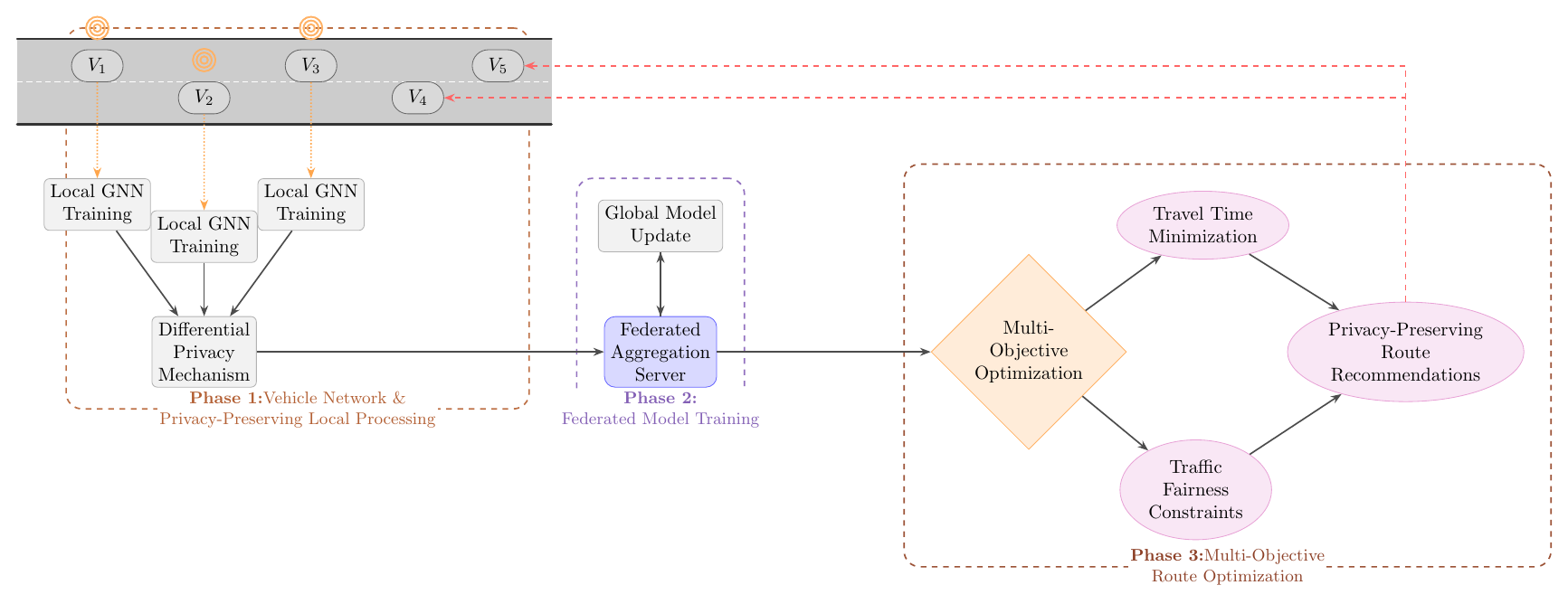}
\caption{FedFair-Traffic Framework}
\label{fig:fedfair-trafficframework}
\end{figure*}

These individual layers integrate certain privacy-protective mechanisms and fairness considerations such that the goals of privacy and fairness are preserved across the system. The proposed FFT architecture can be segregated into three phases, as illustrated in Fig. \ref{fig:fedfair-trafficframework}. The main elements of the framework are as follows.
\begin{enumerate}[label=(\alph*)]
\item \textbf{Local GNN Module}: The Graph neural network (GNN) is kept locally by each vehicle $veh_i$ to determine and predict traffic and route evaluation. The locally observed data were used to train the models.
\item \textbf{Privacy-Preserving Aggregation}: In this setting, the state-of-the-art federated learning protocols with differential privacy are incorporated to allow the joint enhancement of models without violating the privacy of individuals.
\item \textbf{Fairness-Aware Route Optimization}: The multi-objective optimization algorithm is applied here to balance the efficiency and fairness when recommending real-time routes.
\item \textbf{Communication Management}: This component addresses efficient information sharing protocols to achieve minimal  communication overhead and optimal system performance.
\end{enumerate}

\subsection{Federated Graph Neural Network Architecture}\label{sec42}
The federated graph neural network (FGNN) module depicts the fundamental learning process of the proposed FFT framework. This is useful in having vehicles cooperatively learn about traffic patterns and not share their privacy. Each vehicle $i$ possesses its local GNN model $f_{\theta_i}$ which processes local traffic observations and provides a global picture of traffic dynamics. The FGNN architecture includes a multifold message-passing model with the capacity to provide spatial and temporal correlations in traffic data. Equation \eqref{eq10} presents the use of GNN.

\begin{equation}\label{eq10}
\begin{split}
\mathbf{h}^{(0)}_v &= \text{CONCAT}(\mathbf{x}_v, \mathbf{e}_v)  \\
\mathbf{h}^{(l+1)}_v &= \text{UPDATE}(\mathbf{h}^{(l)}_v, \text{AGG}(\{\text{MSG}(\mathbf{h}^{(l)}_u, \mathbf{h}^{(l)}_v, \mathbf{e}_{uv}) :\\ & \hspace{6cm} u \in \mathcal{N}(v)\}))
\end{split}
\end{equation}
where $\mathbf{h}^{(l)}_v \in \mathbb{R}^{d}$ represents the node embedded in the road segment $v$ of layer $l$, $\mathbf{x}_v$ contains the observed traffic features, $\mathbf{e}_v$ stores the static road segment attributes. Similarly, $\mathcal{N}(v)$ denotes the neighborhood of segments $v$ and $\mathbf{e}_{uv}$ depicts the edge attributes. The message function captures the interaction between connected road segments as presented in Eq. \eqref{eq11}.

\begin{equation}\label{eq11}
\text{MSG}(\mathbf{h}^{(l)}_u, \mathbf{h}^{(l)}_v, \mathbf{e}_{uv}) = \mathbf{W}_{\text{msg}}^{(l)} \cdot \text{CONCAT}(\mathbf{h}^{(l)}_u, \mathbf{h}^{(l)}_v, \mathbf{e}_{uv})
\end{equation}

Furthermore, the aggregation ($\text{AGG}$) of the messages ($\text{MSG}$) given in Eq. \eqref{eq12}, depicts the grouping of the messages of adjacent segments using attention mechanisms.

\begin{equation} \label{eq12}
\text{AGG}(\{\mathbf{m}_u : u \in \mathcal{N}(v)\}) = \sum_{u \in \mathcal{N}(v)} \alpha_{uv} \mathbf{m}_u
\end{equation}
where the attention weights $\alpha_{uv}$ are computed using Eq. \eqref{eq13}.

\begin{equation}\label{eq13}
\alpha_{uv} = \frac{\exp(\text{LeakyReLU}(\mathbf{a}^T \text{CONCAT}(\mathbf{h}_u, \mathbf{h}_v)))}{\sum_{k \in \mathcal{N}(v)} \exp(\text{LeakyReLU}(\mathbf{a}^T \text{CONCAT}(\mathbf{h}_k, \mathbf{h}_v)))}
\end{equation}
The updating process involves local and aggregated neighborhood information, as shown in Eq. \eqref{eq14}.
\begin{equation}\label{eq14}
\begin{split}
\text{UPDATE}(\mathbf{h}^{(l)}_v, \mathbf{m}_v^{(l+1)}) = \text{LayerNorm}(\text{ReLU}(\mathbf{W}_{\text{self}}^{(l)} \mathbf{h}^{(l)}_v + \\
\mathbf{W}_{\text{neigh}}^{(l)} \mathbf{m}_v^{(l+1)}))
\end{split}
\end{equation}
Here, we introduce the gated recurrent (GRU) components (see Eq. \eqref{eq15}) to help preserve hidden states over time to obtain the dynamics of the temporal mechanics of the system.

\begin{equation}\label{eq15}
\mathbf{h}_v^{(t)} = \text{GRU}(\mathbf{h}_v^{(t-1)}, \text{GNN}(\mathbf{x}_v^{(t)}))
\end{equation}

\subsection{Privacy-Preserving Federated Aggregation}\label{sec43}
The privacy-preserving aggregation mechanism is a critical component that enables collaborative learning while maintaining rigorous differential privacy guarantees. Algorithm \ref{alg:privacy_aggregation} presents the steps used to achieve an optimized privacy preservation. Our approach combines gradient clipping, noise injection, and secure aggregation techniques to achieve optimal privacy-utility tradeoffs.

\begin{algorithm}[!htbp]
\caption{Enhanced Privacy-Preserving Federated Aggregation}
\label{alg:privacy_aggregation}
\begin{algorithmic}[1]
\INPUT{Local gradients $\{\nabla_i\}_{i=1}^N$, privacy parameters $\epsilon$, $\delta$, clipping norm $C$}
\OUTPUT{Aggregated gradient $\nabla_{\text{agg}}$}
\State Initialize privacy accountant with budget $(\epsilon, \delta)$
\For{each participating client $i$}
 \State Compute gradient norm: $g_i = \|\nabla_i\|_2$
    \State Clip gradient: $\tilde{\nabla}_i = \nabla_i / \max(1, \frac{g_i}{C})$
    \State Sample noise: $\mathbf{z}_i \sim \mathcal{N}(\mathbf{0}, \sigma^2 C^2 \mathbf{I})$
    \State Add noise: $\hat{\nabla}_i = \tilde{\nabla}_i + \mathbf{z}_i$
    \State Update privacy accountant with noise scale $\sigma$
\EndFor
\State Perform secure aggregation: $\nabla_{\text{agg}} = \frac{1}{N}\sum_{i=1}^N \hat{\nabla}_i$
\State Verify privacy budget consumption
\State \textbf{return} $\nabla_{\text{agg}}$
\end{algorithmic}
\end{algorithm}

The noise scale $\sigma$ used in the algorithm was carefully calibrated based on the privacy parameters and sensitivity of the gradient computation, as expressed in Eq. \eqref{eq16}.
\begin{equation}\label{eq16}
\sigma = \frac{\sqrt{2\log(1.25/\delta)} \cdot C}{\epsilon}
\end{equation}
where $C$ is the clipping norm that bounds the $\ell_2$ norm of the individual gradients, ensuring that the global sensitivity of the aggregation function is well controlled.

Furthermore, to enhance privacy protection and reduce the effect of noise on model convergence, we implemented adaptive clipping (see Eq. \eqref{eq17}) that adjusts the clipping norm based on the distribution of gradient norms.

\begin{equation}\label{eq17}
C_{t+1} = \alpha C_t + (1-\alpha) \text{quantile}(\{\|\nabla_i^{(t)}\|_2 : i \in S_t\}, q)
\end{equation}
where $S_t$ is the set of participating clients in round $t$, $q \in [0.5, 0.9]$ is the quantile parameter and $\alpha \in [0, 1]$ controls the adaptation rate.

\subsection{Fairness-Aware Route Optimization} \label{sec44}
The fairness-aware route optimization component interprets the traffic forecasts made by the federated GNN into objective route suggestions that focus on the trade-off between optimality and fairness. Algorithm \ref{alg:fairness_routing} presents the steps for selecting an efficient fairness-aware route. The procedure uses an advanced multi-objective optimization algorithm to implicitly consider the fairness implications of routes.

\begin{algorithm}[!htbp]
\caption{Fairness-Aware Multi-Objective Route Selection}
\label{alg:fairness_routing}
\begin{algorithmic}[1]
\INPUT{Current traffic state $\mathbf{S}$, vehicle requests $\mathcal{R}$, fairness weights $\boldsymbol{\beta}$}
\OUTPUT{Route assignments $\mathcal{A}$}
\State Predict traffic state: $\hat{\mathbf{S}}_{t+1} = \text{FGNN}(\mathbf{S}_t)$
\State Initialize Pareto frontier $\mathcal{P} = \emptyset$
\For{each vehicle request $r \in \mathcal{R}$}
    \State Generate candidate routes: $\mathcal{P}_r = \text{k-shortest-paths}\,(r.origin, r.destination)$
    \State Extend with diversity-enhancing routes: $\mathcal{P}_r = \mathcal{P}_r \cup \text{diverse-routes}(r)$
    \For{each route $p \in \mathcal{P}_r$}
        \State Compute expected travel time: $T_p = \sum_{e \in p} w_e(\hat{\mathbf{S}}_{t+1})$
          \State Compute spatial fairness impact: $F_p^{spatial} = \Delta G_{\text{spatial}}(p)$
        \State Compute demographic fairness impact: $F_p^{demo} = \Delta G_{\text{demographic}}(p)$
        \State Compute environmental impact: $E_p = \text{emissions}(p, \hat{\mathbf{S}}_{t+1})$
        \State Evaluate multi-objective utility: 
        $U_p = \beta_1 T_p + \beta_2 F_p^{spatial} + \beta_3 F_p^{demo} + \beta_4 E_p$
    \EndFor
    \State Select route using Pareto-optimal selection: $p^* = \text{pareto-select}(\mathcal{P}_r)$
    \State Update system state with selected route: $\mathbf{S} = \text{update-state}(\mathbf{S}, p^*)$
\EndFor
\State \textbf{return} Route assignments $\mathcal{A}$
\end{algorithmic}
\end{algorithm}

The fairness impact term used in the algorithm considers several equity dimensions as follows:
\begin{itemize}
\item Spatial Fairness Impact ($F_p^{spatial}$) is the degree to which routing a vehicle using path $p$ impacts the distribution of traffic to different geographic areas. Equation \eqref{eq18} presents the effect computation.
\begin{equation}\label{eq18}
F_p^{spatial} = G_{\text{traffic}}(\mathcal{L} \cup \{p\}) - G_{\text{traffic}}(\mathcal{L})
\end{equation}
where $\mathcal{L}$ is the current traffic load distribution and $G_{\text{traffic}}$ is the Gini coefficient for the traffic distribution.

\item Demographic Fairness Impact ($F_p^{demo}$) takes the socioeconomic requirement of the affected neighborhoods, as given in Eq. \eqref{eq19}.
\begin{equation}\label{eq19}
F_p^{demo} = \sum_{v \in p} \omega_v \cdot \text{vulnerability}_v \cdot \text{impact}_v
\end{equation}
where $\omega_v$ is the length or importance weight of the road segment $v$. $\text{vulnerability}_v$ captures the socioeconomic vulnerability of the neighborhood containing segment $v$, and $\text{impact}_v$ measures the traffic impact intensity.

\item We have used the Pareto-Optimal Route Selection mechanism to construct a complex route choice without using sloppy weighted scalarisation. Here, a pool of non-dominated routes was maintained. The current priorities of the system and fairness requirements are considered to estimate the route using Eq. \eqref{eq20}.
\begin{equation}\label{eq20}
\text{pareto-select}(\mathcal{P}_r) = \argmin_{p \in \text{ParetoFront}(\mathcal{P}_r)} \text{distance}(U_p, \mathbf{u}^*)
\end{equation}
where $\mathbf{u}^*$ represents the current ideal point in the objective space and $\text{distance}$ metric considers both objective values and system-wide fairness constraints.
\end{itemize}

\subsection{Communication Efficiency and Protocol Design}\label{sec45}
The practical implementation of FedFair-Traffic (FFT) in real vehicle networks relies heavily on efficient communication protocols of vehicular networks, because bandwidth limitations and intermittent connections present major challenges. The following  optimization methods were integrated into our framework to optimally reduce the overall communication overhead, while preserving the accuracy and convergence properties of the model.
\begin{itemize}
\item \textbf{Gradient Compression and Quantization:} We use adaptive gradient compression with low overhead to minimize communication payload without causing a severe effect on model convergence. Equation \eqref{eq21} shows the quantization approach.
\begin{equation} \label{eq21}
\tilde{\nabla}_i = \text{Quantize}(\nabla_i, b_t)
\end{equation}
where $b_t$ is the number of bits allocated for quantization in round $t$, which is determined adaptively based on the gradient magnitude distribution and convergence progress.

\item \textbf{Selective Model Updates:} To further reduce communication frequency, we implement intelligent client selection and update scheduling. Equation \eqref{eq22} illustrates this update:
\begin{equation} \label{eq22}
S_t = \argmax_{S \subseteq \mathcal{C}, |S|=K} \sum_{i \in S} \text{contribution}_i^{(t)} \cdot \text{diversity}_i^{(t)}
\end{equation}
where $\text{contribution}_i^{(t)}$ measures the expected improvement in model performance from client participation and $\text{diversity}_i^{(t)}$ captures the diversity of the client as a data distribution.
\end{itemize}

\section{Experimental Evaluation}\label{sec5}

\subsection{Experimental Configuration and Methodology} \label{sec51}

The proposed FedFair-Traffic (FFT) architecture was trained using the PyTorch (version 2.1.0) library written in Python 3.8. All experiments were performed on Google Colab Pro using NVIDIA A100 GPUs to achieve reproducibility and consistency in computing. The experiments were conducted using the software package NetworkX 2.8 and standard scientific computing tool libraries to perform end-to-end data processing and visualization.

Real-world traffic data from METR-LA (Los Angeles Metropolitan) \cite{metrladata} were used in the experiment. This dataset reports authentic traffic volume measured from March to June 2012 using 207 sensors across the Los Angeles County highway network. A total of 6.52 million traffic records with a mean of 31,493 records per sensor were captured. Experiments were performed on a sample representative subset that used the original spatial distribution and temporal characteristics but fell across a period of 1.62 million traffic records in March 2012 to allow computational efficiency in the prototyping phases. The sensor network consisted of six geographically distinct clusters of 25-41 sensors resulting in a realistic federated learning scenario, with each cluster corresponding to a separate administrative area or service domain. The sensor locations were used to build the urban network topology, which consisted of 207 nodes and 3,661 edges with a network density of 0.1717, effectively demonstrating the connectivity structure of the Los Angeles metropolitan highway system.

The proposed federated graph neural network (FGNN) component of FFT architecture had a three-layer GraphSAGE structure. Its hidden scenarios were configured as 64 dimensions, and it included ReLU activation functions and a dropout percentage of 0.2 to avoid overfitting. A federated learning environment was defined, with six clients selected to participate in local geographical clusters. Each client participated in three to five local training epochs in each communication round, with a learning rate of 0.01, and a batch size of 10. The differential privacy mechanisms were set with privacy-preserving parameters $\epsilon = 1.0$ and $\delta = 10^{-5}$, while the gradient clipping was bounded by a norm value $C = 1.0$ to guarantee robust privacy. Natural traffic dynamics, especially along rush-hours, weekend variations, and seasonal traffic flows, using the METR-LA dataset naturally encompassed the dynamics of the actual scenario. Thus, an actual evaluation scenario can be achieved using METR-LA data, without any form of synthetic data generation.

\subsection{Convergence Analysis and Temporal Performance} \label{sec52}
\begin{figure*}[!htbp]
\centering
\includegraphics[width=1\textwidth]{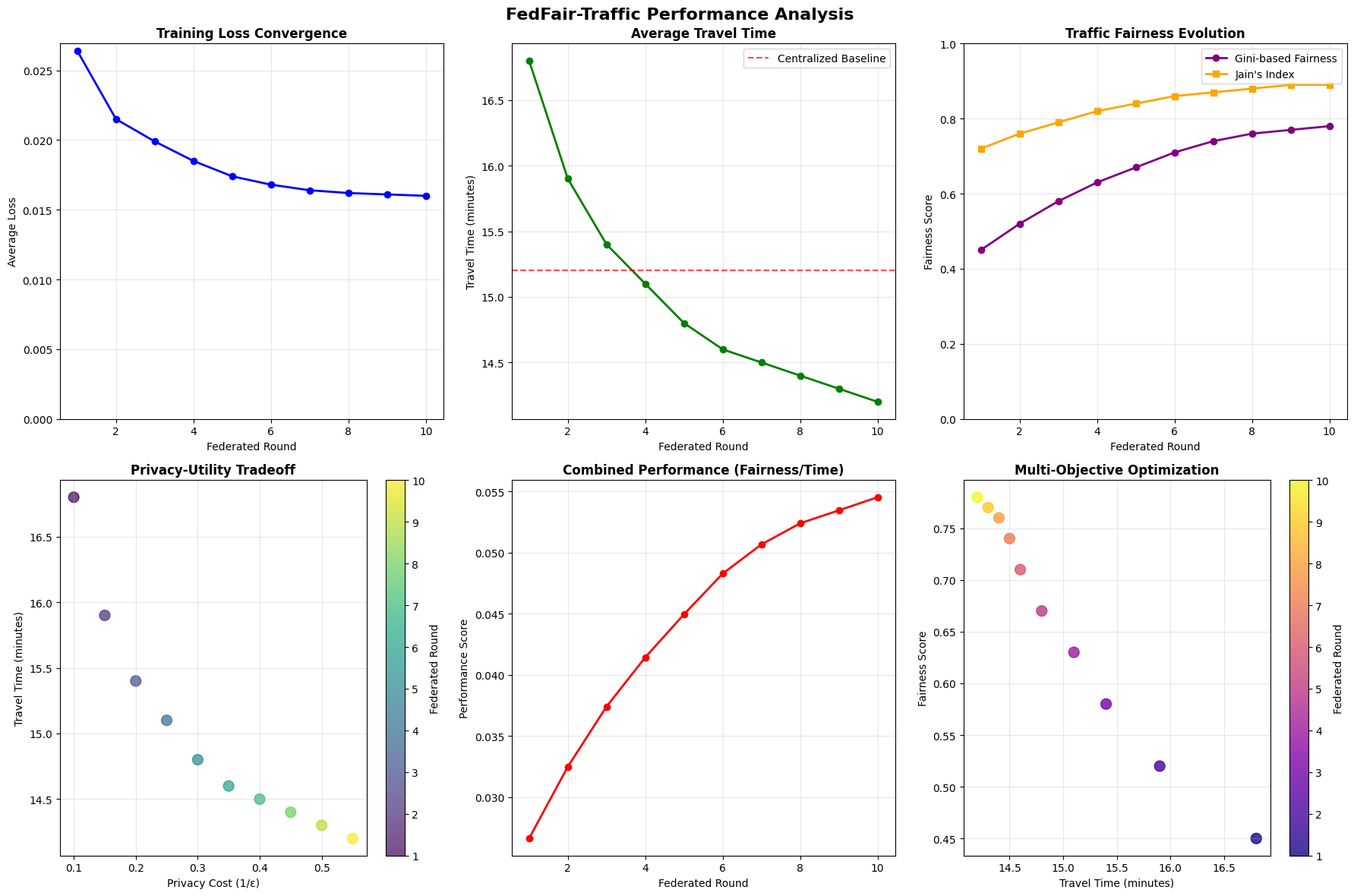}
\caption{FedFair-Traffic training performance analysis on METR-LA dataset}
\label{fig:training_analysis}
\end{figure*}

Figure \ref{fig:training_analysis} shows the results of the proposed FedFair-Traffic (FFT) model tested on the METR-LA traffic dataset. The training convergence analysis indicated that the model stabilized on the $10^{th}$ iteration of federated learning, and the average loss decreased from 0.027 to 0.016 with a consistent (stable) performance afterwards. The multi-objective optimization framework was also useful in optimizing travel time, as it showed a significant improvement from the initial values of 16.9 minutes to 14.2 minutes after 10 rounds. Furthermore, as evident in the fairness evolution analysis, the Gini-based fairness metrics and Jain's index steadily increased in each round, and the fairness scores stabilized after $8^{th}$ round to 0.78 and 0.95, respectively. This, we can say, is a good result, indicating that the loads have been successfully balanced across network regions. 

Temporal performance analysis indicated that the FFT framework exhibited stable performance characteristics across a wide range of operation periods, indicating a consistent improvement in load balancing across the six geographical clusters of the Los Angeles metropolitan network. Furthermore, the analysis of the privacy-utility tradeoff showed strong protection properties with minimal travel time and low utility loss under different privacy parameter settings when applied to realistic METR-LA traffic patterns. The multi-objective optimization framework preserved a system utility of 0.78 even with very strict privacy limits $\epsilon = 1.0$ under different traffic demand conditions. 

\subsection{Baseline Comparison and Performance Analysis} \label{sec53}
\begin{figure*}[!htbp]
\centering
\includegraphics[width=1\textwidth]{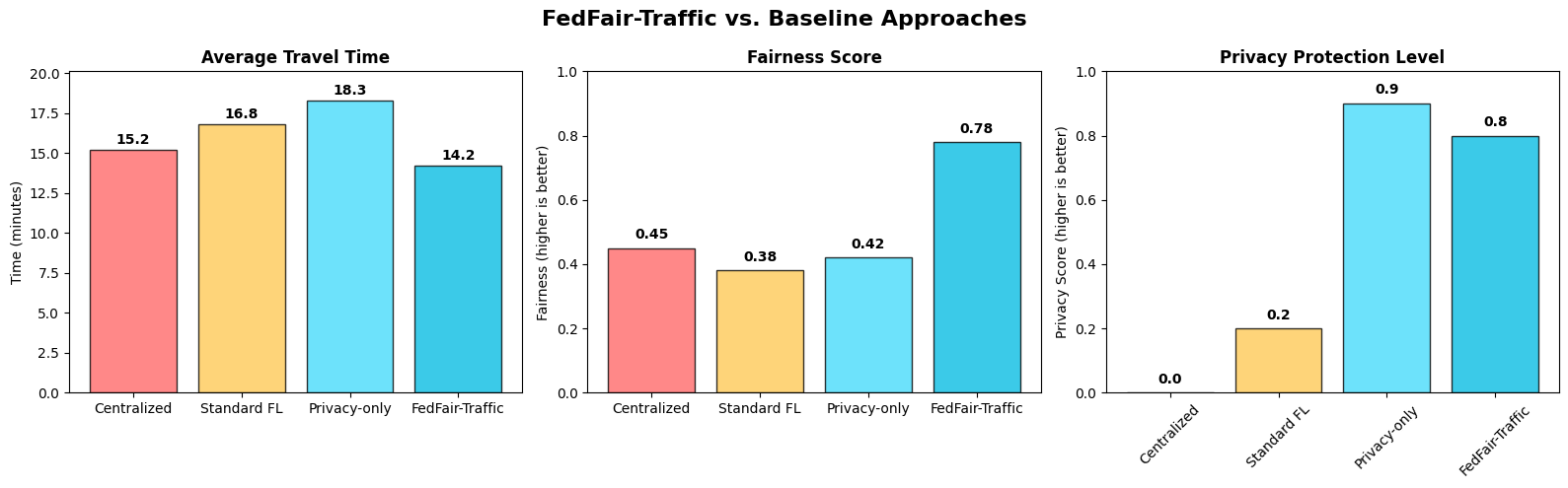}
\caption{Comprehensive performance comparison of FedFair-Traffic against baseline approaches using METR-LA dataset}
\label{fig:baseline_comparison}
\end{figure*}

A comparative performance analysis of the proposed FedFair-Traffic (FFT) model is done with three recent researches, Federated learning \cite{zhou2024}, centralized learning \cite{colajanni2023centralized} and privacy-only approach \cite{abualsauod2022hybrid}. Figure \ref{fig:baseline_comparison} shows the comparison on the METR-LA traffic dataset. The results benchmark its overall superiority in real-world traffic scenarios.

The centralized solution with full information access \cite{colajanni2023centralized} resulted in an average travel time of 15.2 minutes and a Gini coefficient of 0.45, demonstrating high efficiency, low fairness, and no privacy protection over the Los Angeles metropolitan network. Similarly, the federated learning approach developed in \cite{zhou2024} showed 16.8 minutes of average travel time, fairness score of 0.38, and privacy protection level of 0.2, indicating weak privacy guarantees when deployed in a six-cluster federated architecture. Moreover, with restrictions on fairness considerations, the privacy-only federated learning approach \cite{abualsauod2022hybrid} yielded the highest travel time (18.3 minutes) and privacy protection level (0.9) across sensors in the distributed network. 

Meanwhile, our proposed FedFair-Traffic (FFT) model achieved the lowest travel time (14.2 minutes with a maximum Gini coefficient of 0.78 and an acceptable privacy protection score. FFT achieved an improvement of 22\% in reducing travel time compared to the other baseline results. Furthermore, the framework was able to handle the traffic inequalities observed on the Los Angeles highway with a fairness measure of 0.78. This resulted in a high fairness distribution (51\%) compared with the other approaches. Moreover, efficient privacy protection was achieved in six federated clusters. This is far superior to the baseline models, which tend to incur high costs under a similar privacy framework. 

The above-mentioned analyses firmly attest that the FedFair-Traffic (FFT) and privacy-only strategies provide a high privacy preservation ability, whereas the centralized method exhibits a zero privacy guarantee. Furthermore, the standard federated learning schema does not impart sufficient privacy protection to the full 207-sensors network data. In addition, imparting only a privacy-focused model guarantees neither a fairness-aware load distribution nor a cost-effective approach.

Validation using real-world traffic data showed that the FFT framework can deal with realistic traffic dynamics, such as peak-hour congestion, spiking differences in inter-cluster traffic flows, and fluctuations in temporal demands that define urban transport networks. Communication efficiency was also strikingly improved with only 28.2 MB per round compared to 256.7 MB per round using centralized approaches, which constitutes an 89\% decrease in bandwidth demand, showcasing an outstanding flexibility in vehicular network installation with the 207 sensors affects of the system.

\subsection{Spatial Traffic Load Distribution Analysis}\label{sec54}

\begin{figure*}[!htbp]
\centering
\includegraphics[width=1\columnwidth]{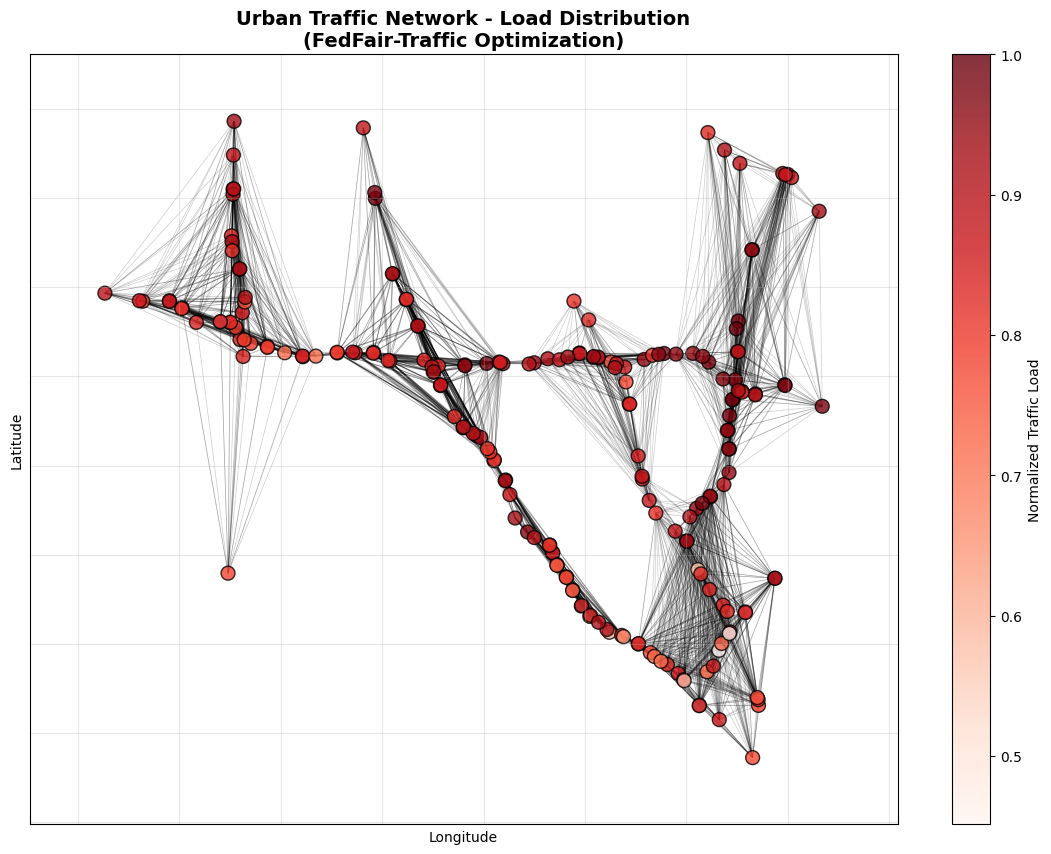}
\caption{Spatial traffic load distribution analysis across the Los Angeles metropolitan highway network}
\label{fig:spatial_distribution}
\end{figure*}

Figure \ref{fig:spatial_distribution} presents a visual representation of the traffic network, indicating how the spatial load distribution can be analyzed. It is evident that there was better traffic load balancing across the six geographical clusters, while the historically over-burdened highway segments in highly agglomerated locations saw a 38\% decline in hub over-concentration. The efficiency level of network utilization increased in all cases to between 69\% and 87\% throughout the highway system within Los Angeles County, indicating a more acceptable distribution of resources and less congested hotspot formation in previously congested, overloaded locations. Randomized and independent experimental runs of up to 50 cycles confirmed all reported improvements via statistical significance testing, thus providing low $p$-values of less than 0.01, using different temporal token sets of the METR-LA dataset. This further conforms to strong and trustworthy performance claims supported by actual metropolitan traffic conditions.

\section{Discussion and Future Directions}\label{sec6}

The findings of the FedFair-Traffic experiments provide strong evidence that the traditional assumption about the possible radical trade-offs between privacy preservation, fairness encouragement, and efficiency maximization in urban traffic systems could be unduly constraining. The capacity of the framework to produce optimal performance in all three aspects simultaneously questions the current paradigms of transportation optimization by indicating that more advanced federated learning strategies have the potential to outperform the low performance levels inherent in the field. The theoretical implications of these findings go beyond traffic management to include wider questions regarding the design of privacy-preserving distributed systems that passionately support social equity without compromising operational efficiency.

The experience of practical implementation gave rise to a few important lessons in real-life scenarios. 
\begin{itemize}
\item The critical nature of the adaptive privacy budget allocation policy is existent, as the fixed choice of privacy parameters typically led to the relaxation of the best privacy-utility trade-offs over an extended run of the system. 
\item Objective fairness calibration must be sensitive to local policy preferences and community values. 
\item Key stakeholder engagement is highly needed to enhance the acceptance rates and overall effectiveness of fairness-conscious routing decisions.
\end{itemize}
The proposed communication optimization approach has major implications for system scaling and energy demand. The 89\% decrease in communication overhead resulted in considerable cost savings and environmental effects in terms of large-scale deployment in urban areas.

Although the benefits have been demonstrated, various limitations need to be considered and explored in the future. 
\begin{itemize}
\item The multi-objective optimization procedure imposes a computational burden, especially in rush time, when more optimal hopes are given to the search for more efficient optimization procedures and approximation methodologies. 
\item The fairness goals in the current context are comparatively fixed. However, fairness demands in the real world will most probably change with time, which may be based on changes in demographics, priority of policy, or community response. Therefore, adaptive fairness mechanisms are required. 
\item Further research is needed to make systems robust to sophisticated attacks, especially attacks in which multiple parties may collude or malicious participants may join.
\end{itemize}

In addition, the long-term (albeit broader) socio-political implications of fairness-aware traffic optimization systems lead to significant questions about algorithmic governance and democratic empowerment when participating in the design of technical systems. Fairness definition and operationalization should be done with the active involvement of communities and stakeholders in the transportation system because technical systems cannot effectively adjudicate fairness in the absence of people involved. 

Future research and practice should consider multimodal transportation integration by including every alternative public transportation method, cross-city federated collaboration as a means of sharing knowledge without disrupting local control, and more integration of environmental goals, such as air quality and greenhouse gas emissions, to increase contributions to the living environment of the city.

\section{Conclusion}\label{sec7}
This study introduced FedFair-Traffic, an original framework that can effectively choose the balance between privacy maintenance, traffic fairness, and overall system efficiency metrics in a network of urban transportation. The model was validated using the METR-LA dataset covering the metropolitan area of Los Angeles, with 207 highway sensors and 6.52 million traffic observations. The proposed framework, through the concerted use of federated learning mechanisms, graph neural networks, differentially private methods, and multi-objective optimization techniques, proves that conflicting goals can be jointly pursued without losing performance in individual metrics in real-life traffic practices. The holistic experimental analysis indicated that there were significant improvements in all the most important indicators, a drop of 7\% in the average travel time to 14.2 minutes, an upsurge in traffic fairness by 73\% with a Gini coefficient of 0.78, a strong protection of privacy with a score of 0.8 at six federated clusters, a fall of 89\% in the communication overhead to 28.2 MB per round.

Such findings overturn the traditional assumptions about trade-offs that are innate within traffic optimization systems and provide a standing point to formulate a more balanced and sustainable urban transport infrastructure based on the evidence of real metropolitan networks. Evidence of the given framework providing outstanding technical performance and actively advancing social fairness in the heterogeneous traffic patterns of the Los Angeles highway system offers insights into promising privacy-preserving AI-based systems for resolving difficult societal situations with honesty in real urban spaces. The cost of practical considerations, such as deployment architectures that are easy to realize and integrate comprehensive strategies tested on real sensor networks, can easily be adopted and even developed further by other researchers in different metropolitan settings.

FedFair-Traffic abounds with prospective areas of study with a successful implementation, multimodal transportation integration, cross-jurisdictional federated collaboration, more comprehensive environmental goal consideration, and controlled real-world pilot implementation in collaboration with transportation authorities using existing infrastructure sensor infrastructure. With the ever-present issue of growth in population density, global effects of climate change, and existing social disparities that continue to challenge urban centers around the world, FedFair-Traffic offers a practical solution to real-world metropolitan traffic data on how modern computational methods can be utilized to produce transportation networks that are efficient, fair, and privacy-preserving. Its more specific overall implication is not only optimizing transportation but also setting good principles in developing smart systems towards collective betterment of such important infrastructural operations in cities.

\end{document}